\documentclass[10pt,twocolumn,letterpaper]{article}

\usepackage{cvpr}
\usepackage{times}
\usepackage{epsfig}
\usepackage{graphicx}
\usepackage{amsmath}
\usepackage{amssymb}
\usepackage{enumitem}
\usepackage{parskip}

\usepackage{color}
\definecolor{turquoise}{cmyk}{0.65,0,0.1,0.3}
\definecolor{purple}{rgb}{0.65,0,0.65}
\definecolor{dark_green}{rgb}{0, 0.5, 0}
\definecolor{orange}{rgb}{0.8, 0.6, 0.2}
\definecolor{red}{rgb}{0.8, 0.2, 0.2}
\definecolor{darkred}{rgb}{0.6, 0.1, 0.05}
\definecolor{blueish}{rgb}{0.0, 0.3, .6}
\definecolor{light_gray}{rgb}{0.7, 0.7, .7}
\definecolor{pink}{rgb}{1, 0, 1}
\definecolor{greyblue}{rgb}{0.25, 0.25, 1}

\renewcommand{\paragraph}[1]{{\textbf{#1.}}}
\newcommand{\Figure}[1]{Figure~\ref{fig:#1}}

\newcommand{\eq}[1]{(\ref{eq:#1})}

\newcommand{\Sec}[1]{Sec.~\ref{sec:#1}}

\let\oldin\in
\renewcommand{\in}{\!\oldin\!}

\usepackage{comment}
\specialcomment{DRAFT}{\color{red}}{\color{black}}
\newcommand{\render}{\mathcal{R}}
\newcommand{\select}{\mathcal{S}}
\newcommand{\warppars}{{\omega}}
\newcommand{\warp}{\mathcal{W}}
\newcommand{\blendpars}{{\beta}}
\newcommand{\blend}{\mathcal{B}}

\newcommand{\imageoutput}{I_{out}}
\newcommand{\imageinput}{\bar{I}}
\newcommand{\imagecloud}{I_\text{cloud}}
\newcommand{\inputcloud}{\bar{\mathbf{P}}}
\newcommand{\cloud}{{\mathbf{P}}}
\newcommand{\inputview}{\bar{v}}
\newcommand{\imagenew}{{I}_\text{cloud}}
\newcommand{\imagecalib}{\bar{I}_\text{calib}}
\newcommand{\imagewarp}{{I}_\text{warp}}
\newcommand{\imagewarpsilho}{{I}^\bullet_\text{warp}}

\newcommand{\imagegt}{I_\text{gt}}

\newcommand{\semanticgt}{{I}^\bullet_\text{gt}}

\newcommand{\bodywarpsilho}{{I}^\bullet_\text{part,p}}
\newcommand{\bgwarpsilho}{{I}^\bullet_\text{bg}}
\newcommand{\bgwarpsilhogt}{{I}^\bullet_\text{bg,gt}}
\newcommand{\warpedtexture}{\bar{I}_\text{warp,p}}
\newcommand{\warpedmask}{\bar{I}^\bullet_\text{part warp}}

\newcommand{\normapenew}{{N}}
\newcommand{\viewnew}{{v}}
\newcommand{\confidence}{{c}}
\newcommand{\posecalib}{\bar{\kappa}_\text{calib}}
\newcommand{\posenew}{{\kappa}}
\newcommand{\pose}{\kappa}
\newcommand{\Input}{\bar}
\newcommand{\poseinput}{\Input\kappa}

\newcommand{\R}{\mathbb{R}} 
\newcommand{\optcenter}{\mathbf{o}}
\newcommand{\focallength}{\mathbf{f}}

\usepackage[pagebackref=true,breaklinks=true,letterpaper=true,colorlinks,bookmarks=false]{hyperref}

\cvprfinalcopy 

\def\cvprPaperID{4204} 

\ifcvprfinal\fi
\begin{document}

\title{Volumetric Capture of Humans with a Single RGBD Camera via  \\ Semi-Parametric Learning}

\author{Rohit Pandey, Anastasia Tkach, Shuoran Yang, Pavel Pidlypenskyi, Jonathan Taylor, \\ Ricardo Martin-Brualla, Andrea Tagliasacchi, George Papandreou, Philip Davidson,  \\ Cem Keskin, Shahram Izadi, Sean Fanello\\
Google Inc.\\
}

\maketitle
\thispagestyle{empty}

\begin{abstract}
Volumetric (4D) performance capture is fundamental for AR/VR content generation.
Whereas previous work in 4D performance capture has shown impressive results in studio settings, the technology is still far from being accessible to a typical consumer who, at best, might own a single RGBD sensor.
Thus, in this work, we propose a method to synthesize free viewpoint renderings using a single RGBD camera.
The key insight is to leverage previously seen ``calibration'' images of a given user to extrapolate what should be rendered in a novel viewpoint from the data available in the sensor.
Given these past observations from multiple viewpoints, and the current RGBD image from a fixed view, we propose an end-to-end framework that fuses both these data sources to generate novel renderings of the performer.
We demonstrate that the method can produce high fidelity images, and handle extreme changes in subject pose and camera viewpoints.
We also show that the system generalizes to performers not seen in the training data.
We run exhaustive experiments demonstrating the effectiveness of the proposed semi-parametric model (i.e. calibration images available to the neural network) compared to other state of the art machine learned solutions.
Further, we compare the method with more traditional pipelines that employ multi-view capture.
We show that our framework is able to achieve compelling results, with substantially less infrastructure than previously required.
\end{abstract}
\begin{figure}[t]
\centering
\includegraphics[width=\columnwidth]{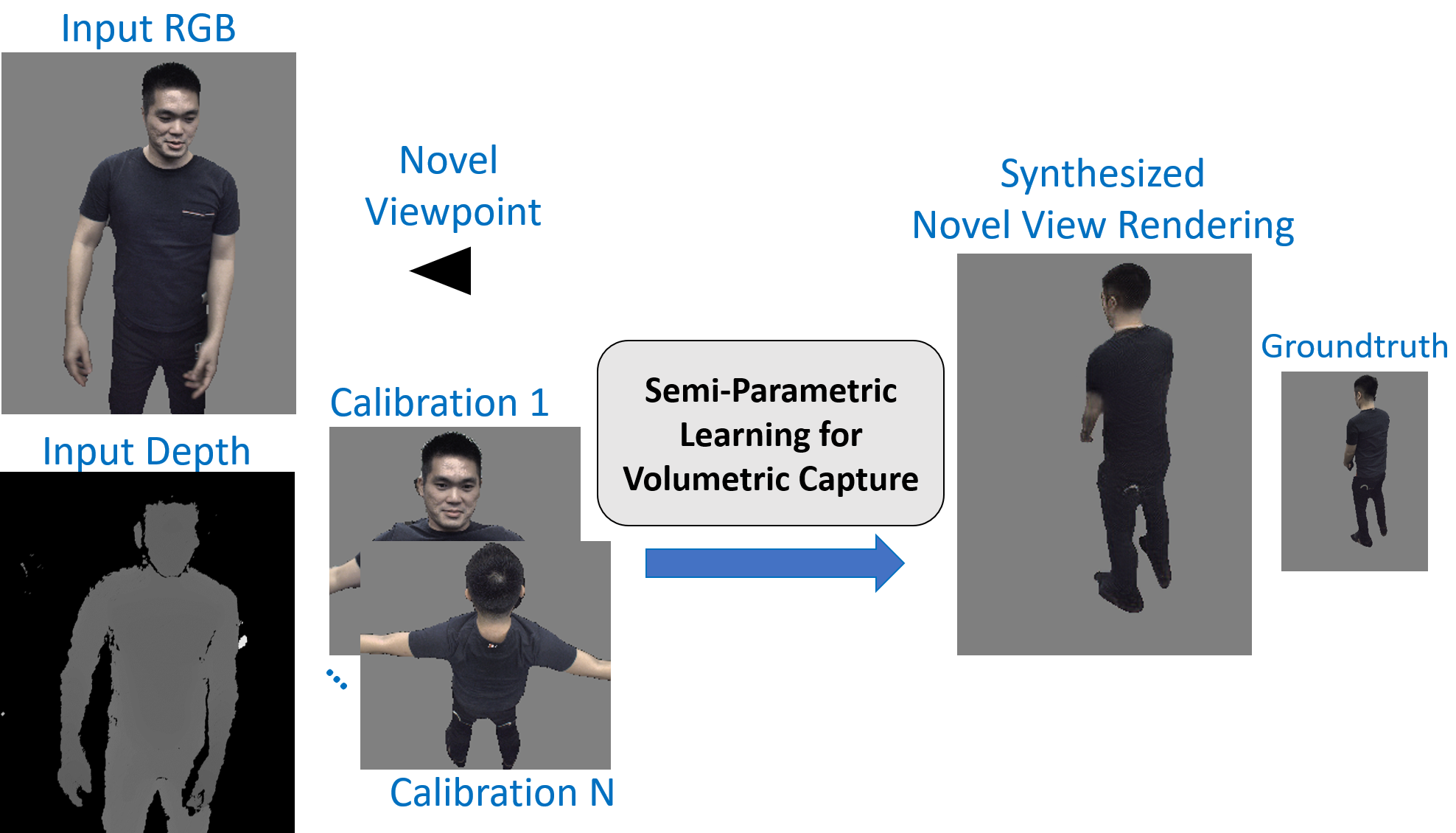}
\caption{
We propose a novel formulation to synthesize volumetric renderings of human from arbitrary viewpoints.
Our system combines previously seen observations of the user (calibration images) with the current RGBD image.
Given an arbitrary camera position we can generate images of the performer handling different user poses and generalizing to unseen subjects.
%
}
\label{fig:teaser}
\vspace{-15pt}
\end{figure}

\section{Introduction}
The rise of Virtual and Augmented Reality has increased the demand for high quality 3D content to create compelling user experiences where the real and virtual world seamlessly blend together.
Object scanning techniques are already available for mobile devices \cite{3dscanner}, and they are already integrated within AR experiences \cite{ARCore}.
{However, neither the industrial nor the research community have yet been able to devise practical solutions to generate high quality volumetric renderings of humans.}

{At the cost of reduced photo-realism, the industry is currently overcoming the issue by leveraging ``cartoon-like'' virtual avatars.}
{On the other end of the spectrum,} complex capture rigs \cite{fvv,prada,Carranza2003FreeViewpointVideo} can be used to generate very high quality volumetric reconstructions.
Some of these methods \cite{lightstage1,lightstage2} are well established, {and lie at the foundation of special effects in many Hollywood productions.}
Despite their success, these systems rely on high-end, costly infrastructure to process the high volume of data that they capture.
The required computational time of several minutes per frame make them  unsuitable for \textit{real-time applications}.
{Another way to capture humans is to extend real-time non-rigid fusion pipelines \cite{dynamicfus,volumedeform,killing,sobolev,twinfusion} to multi-view capture setups \cite{dou16,holoportation,dou17}.}
However, the results still suffer from \textit{distorted geometry}, poor texturing and inaccurate lighting, making it difficult to reach the level of quality required in AR/VR applications~\cite{holoportation}.
Moreover, these methods rely on multi-view capture rigs that require several~($\approx$ 4-8) calibrated RGBD sensors.

{Conversely, our goal is to make the volumetric capture technology accessible through consumer level hardware.} 
Thus, in this paper, we focus on the problem of synthesizing volumetric renderings of humans.
Our goal is to develop a method that leverages recent advances in machine learning to generate 4D videos using \textit{as little infrastructure as possible} -- a single RGBD sensor.
{We show how a semi-parametric model, where the network is provided with calibration images,} can be used to render an image of a novel viewpoint by leveraging the calibration images to extrapolate the partial data the sensor can provide.
Combined with a fully parametric model, this produces the desired rendering from an arbitrary camera viewpoint; see Fig. \ref{fig:teaser}.

%
In summary, our contribution is a new formulation of volumetric capture of humans that employs a single RGBD sensor, and that leverages machine learning for image rendering. Crucially, our pipeline does not require complex infrastructure typically required by 4D video capture setups.

We perform exhaustive comparisons with machine learned, as well as traditional state-of-the-art capture solutions, showing how the proposed system generates compelling results with minimal infrastructure requirements.

\section{Related work}
Capturing humans in 3D is an active research topic in the computer vision, graphics, and machine learning communities.
We categorize related work into three main areas that are representative of the different trends in the literature: \textit{image-based rendering}, \textit{volumetric capture}, and \textit{machine learning solutions}.

\paragraph{Image based rendering}
{Despite their success, most of methods in this class do not infer a full 3D model, but can nonetheless generate renderings from novel viewpoints.
Furthermore, the underlying 3D geometry is typically a proxy, which means they cannot be used in combination with AR/VR where accurate, \textit{metric} reconstructions can enable additional capabilities. 
For example, \cite{DebevecFacade,GortlerLumigraph}, create impressive renderings of humans and objects, 
but with limited viewpoint variation.}
Modern extensions \cite{anderson2016jump,megastereo} produce $360^\circ$ panoramas, but with a fixed camera position.
%
%
The method of Zitnick~\etal~\cite{zitnick04}
infers an underlying geometric model by predicting proxy depth maps, but with a small $30^\circ$ coverage, and the rendering heavily degrades when the interpolated view is far from the original. 
Extensions to these methods \cite{eisemann,casas,volino} have \textit{attempted} to circumvent these problems by introducing an optical flow stage warping the final renderings among different views, but with limited success.

\paragraph{Volumetric capture}
Commercial volumetric reconstruction pipelines employ capture studio setups to reach the highest level of accuracy~\cite{fvv,prada,dou16,dou17,holoportation}.
For instance, the system used in \cite{fvv,prada}, employs more than $100$ IR/RGB cameras, which they use to accurately estimate depth, and then reconstruct 3D geometry~\cite{poisson_surface_reconstruction}.
Non-rigid mesh alignment and further processing is then performed to obtain a temporally consistent atlas for texturing.
Roughly 28 minutes per frame are required to obtain the final 3D mesh.
Currently, this is the state-of-the-art system, and is employed in many AR/VR productions.
Other methods \cite{zollhoefer2014,dynamicfus,dou16,dou17,holoportation,Du2018Montage4D}, further push this technology by using highly customized, high speed RGBD sensors.
{High framerate cameras~\cite{fanello2017ultrastereo,fanello17_hashmatch,sos} can also help make the non-rigid tracking problem more tractable, and compelling volumetric capture can be obtained with just 8 custom RGBD sensors rather than hundreds~\cite{need4speed}.}
%
However these methods still suffer from both geometric and texture aberrations, as demonstrated by Dou~\etal\cite{dou17} and Du~\etal~\cite{Du2018Montage4D}. 

\paragraph{Machine learning techniques}
The problem of generating images of an object from novel viewpoints can also be cast from a machine learning, as opposed to graphics, standpoint.
For instance, Dosovitskiy~\etal\cite{learningchairs} generates re-renderings of chairs from different viewpoints, but the quality of the rendering is low, and the operation is specialized to discrete shape classes. 
%
More recent works \cite{deepiewmorphing,transfgrounded,appearanceflow} try to learn the 2D-3D mapping by employing some notion of 3D geometry, or to encode multiview-stereo constraints directly in the network architecture~\cite{deepstereo}.
As we focus on humans, our research is more closely related to works that attempt to synthesize 2D images of humans~\cite{acm18,balakrishnan_cvpr18,si_2018_CVPR,ma18,ma17,Guler2018DensePose,everybodydance}.
These focus on generating people in unseen poses, but usually from a fixed camera viewpoint (typically frontal) and scale {(not metrically accurate)}.
The coarse-to-fine GANs of \cite{acm18} synthesizes images that are still relatively blurry. Ma~\etal~\cite{ma17} detects pose in the input, which helps to disentangle appearance from pose, resulting in improved sharpness.
Even more complex variants \cite{ma18,si_2018_CVPR} that attempt to disentangle pose from appearance, and foreground from background, still suffer from multiple artifacts, especially in occluded regions.
A dense UV map can also be used as a proxy to re-render the target from a novel viewpoint~\cite{Guler2018DensePose}, but high-frequency details are still not effectively captured.
Of particular relevance is the work by Balakrishnan~\etal\cite{balakrishnan_cvpr18}, where through the identification and transformation of body \textit{parts}  results in much sharper images being generated.
Nonetheless, note how this work solely focuses on \textit{frontal} viewpoints.
 
\paragraph{Our approach}
In direct contrast, our goal is to render a subject in  \textit{unseen poses} and \textit{arbitrary viewpoints}, mimicking the behavior of volumetric capture systems.
The task at hand is much more challenging because it requires disentangling pose, texture, background and viewpoint simultaneously.
This objective has been \textit{partially} achieved by Martin-Brualla~\etal\cite{lookingood} by combining the benefits of geometrical pipelines \cite{dou17} to those of convolutional architectures~\cite{unet}.
However, their work still necessitates a complete mesh being reconstructed from multiple viewpoints.
In contrast, our goal is to achieve the same level of photo-realism from a single RGBD input.
To tackle this, we resort to a semi-parametric approach~\cite{sims}, where a calibration phase is used to acquire frames of the users appearance from a few different viewpoints.
These calibration images are then merged together with the the current view of the user in an end-to-end fashion.
We show that the semi-parametric approach is the key to generating high quality, 2D renderings of people in arbitrary poses and camera viewpoints.

\begin{figure*}[t]
\centering
\includegraphics[width=0.7\linewidth]{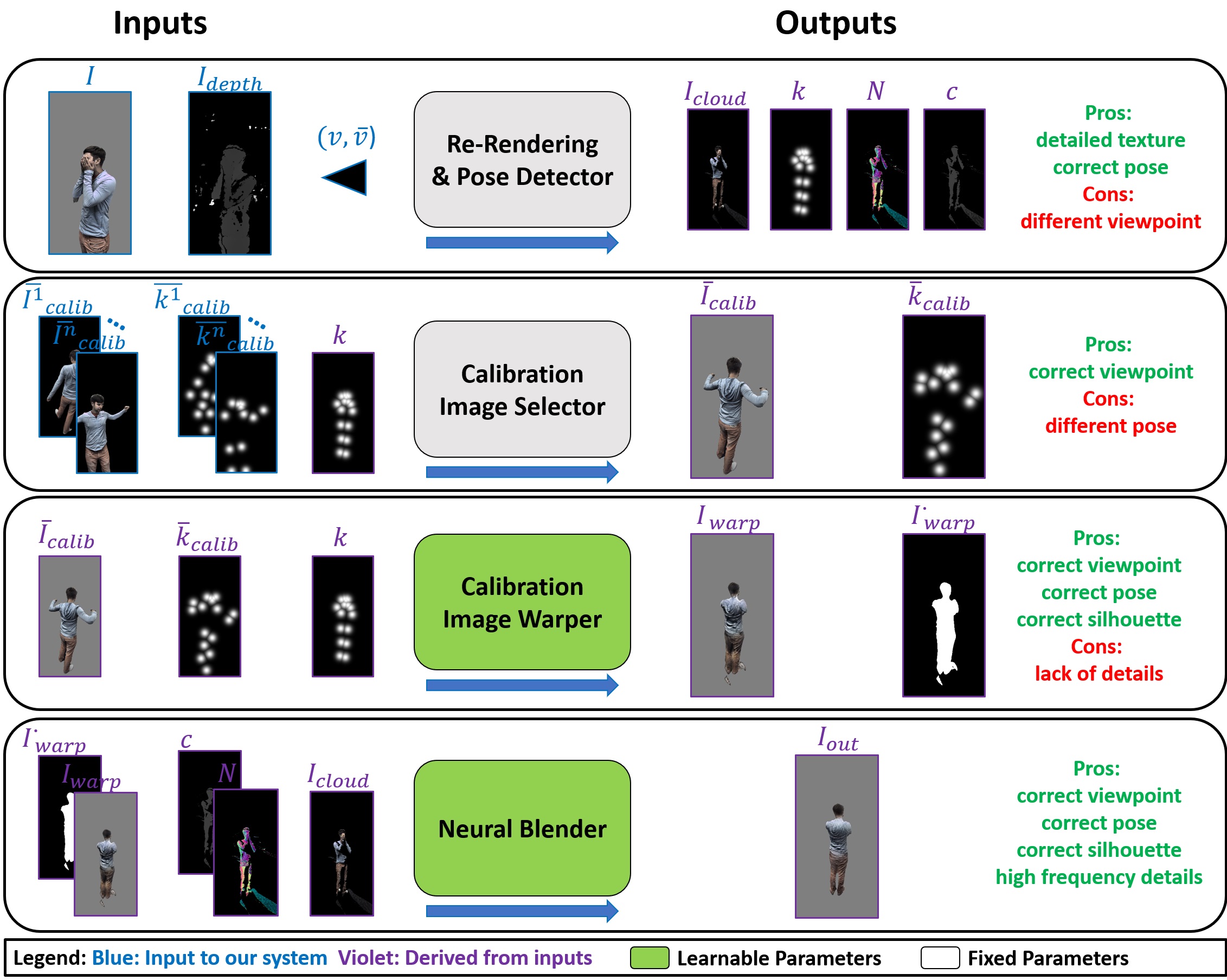}
\caption{
\textbf{Proposed framework} -- 
We take in input the current RGBD image, a novel viewpoint and a collection of images acquired in a calibration stage, which depict the users in different poses observed from several viewpoints.
The \textit{Re-rendering \& pose-detector} projects the texture using depth information and re-project back to the final viewpoint, together with the target pose. We also compute a confidence score of the current observations with respect to the novel viewpoint. This score is encoded in the normal map $\normapenew$ and the confidence $\confidence$.
The \textit{Calibration Image Selector} picks the closest image (in terms of viewpoint) from a previously recorded calibration bank.
The \textit{Calibration Image Warper} tries to align the selected calibration image with the current pose, it also produces a silhouette mask.
The \textit{Neural Blender} combines the information from the warped RGB image, aligned calibration image, silhouette image and viewpoint confidence to recover the final, highly detailed RGB image.
}
\label{fig:framework}
\vspace{-15pt}
\end{figure*}

\section{Proposed Framework}

As illustrated in \Figure{teaser}, our method receives as input:
1)~an RGBD image from a single viewpoint,
2)~a novel camera pose with respect to the current view and
3)~a collection of a few calibration images observing the user in various poses and viewpoints.
As output, it generates a rendered image of the user as observed from the \textit{new} viewpoint. 
Our proposed framework is visualized in \Figure{framework}, and includes the four core components outlined below.

\begin{description}[noitemsep,leftmargin=0em]
\item[Re-rendering \& Pose Detector:]
from the RGBD image $\imageinput$ captured from a camera $\inputview$, we re-render the colored depthmap from the new camera viewpoint $\viewnew$ to generate an image $\imagenew$, as well as its approximate normal map $\normapenew$.
Note we only re-render the foreground of the image, by employing a fast background subtraction method based on depth and RGB as described in~\cite{fanello17_hashmatch}.
We also estimate the pose $\posenew$ of the user, i.e. keypoints, in the coordinate frame of $\viewnew$, as well as a scalar confidence $\confidence$, measuring the divergence between the camera viewpoints: 
\begin{equation}
\imagecloud,\posenew,\normapenew,\confidence  = \render{}(\imageinput,\inputview,\viewnew).
\label{eq:render}
\end{equation}

\item[Calibration Image Selector:]
from the collection of calibration RGBD images and poses $\{\imagecalib^n, \posecalib^n\}$, we select one that best resembles the target pose $\posenew$ in the viewpoint $\viewnew$:
\begin{equation}
\imagecalib, \posecalib = \select{}(\{\imagecalib^n, \posecalib^n\},\posenew).
\label{eq:select}
\end{equation}

\item[Calibration Image Warper:]
given the selected calibration image $\imagecalib$ and the user's pose $\posecalib$, a neural network $\warp$ with learnable parameters $\warppars$ \textit{warps} this image into the desired pose $\posenew$, while simultaneously producing the silhouette mask $\imagewarpsilho$ of the subject in the new pose:
\begin{equation}
\imagewarp, \imagewarpsilho = \warp_\warppars{}(\imagecalib,\posecalib,\posenew).
\label{eq:warper}
\end{equation}
 
\item[Neural Blender:]
finally, we \textit{blend} the information captured by the traditional re-rendering in \eq{render} to the warped calibration image \eq{warper} to produce our final image $\imageoutput$:
\begin{equation}
\imageoutput = \blend_{\blendpars}(\imagenew, \imagewarp, \imagewarpsilho, \normapenew, \confidence).
\label{eq:blender}
\end{equation}
\end{description}
Note that while \eq{render} and \eq{select} are not learnable, they extract quantities that express the geometric structure of the problem.
Conversely, both warper~\eq{warper} and~\eq{blender} are differentiable and trained end-to-end where the loss is the weighted sum between warper $\mathcal{L}_\text{warper}$ and blender $\mathcal{L}_\text{blender}$ losses. 
The weights $\omega_\text{warper}$ and $\omega_\text{blender}$ are chosen to ensure similar contributions between the two.
We now describe each component in details, motivating the design choices we took.

\subsection{Re-rendering \& Pose Detector}
We assume that camera intrinsic parameters (optical center $\optcenter$ and focal length $\focallength$) are known and thus the function $\Pi^{-1}(\mathbf{p},z | \optcenter, \focallength): \R{}^3 \mapsto \R{}^3$ maps a 2D pixel $\mathbf{p}=(x,y)$ with associated depth $z$ to a 3D point in the \textit{local} camera coordinate frame. 

\paragraph{Rendering $\rightarrow \imagecloud$}
\newcommand{\cloudsize}{M}
Via the function $\Pi^{-1}$, we first convert the depth channel of $\imageinput$ into a point cloud of size $\cloudsize$ in matrix form as $\inputcloud \in \R^{4 \times \cloudsize}$.
We then rotate and translate this point cloud into the novel viewpoint coordinate frame as $\cloud = \mathbf{T} \inputcloud$, where $\mathbf{T} \in \mathbb{R}^{4 \times 4}$ is the homogeneous transformation representing the relative transformation between $\inputview$ and $\viewnew$. 
We render $\cloud$ to a 2D image $\imagecloud$ in OpenGL by splatting each point with a $3 \times 3$ kernel to reduce re-sampling artifacts.
Note that when input and novel camera viewpoints are close, i.e. $\inputview \sim \viewnew$, then $\imageoutput \sim \imagecloud$, while when $\inputview \nsim \viewnew$ then $\imagecloud$ would mostly contain unusable information.


\paragraph{Pose detection $\rightarrow \posenew$}
We also infer the pose of the user by computing 2D keypoints  $\poseinput_\text{2D} = \mathcal{K}_\gamma(\imageinput)$ using the method of Papandre~\etal\cite{posenet} where $\mathcal{K}$ is a pre-trained feed-forward network.
We then lift 2D keypoints to their 3D counterparts $\poseinput$ by employing the depth channel of $\imageinput$ and, as before, transform them in the camera coordinate frame $\viewnew$ as $\posenew$.
We extrapolate missing keypoints when possible relying on the rigidity of the limbs, torso, face, otherwise we simply discard the frame.
Finally, in order to feed the keypoints $\posenew$ to the networks in~\eq{warper} and~\eq{blender} following the strategy in \cite{balakrishnan_cvpr18}: we encode each point in an image channel (for a total of $17$ channels) as a Gaussian centered around the point with a fixed variance.
We tried other representations, such as the one used in \cite{si_2018_CVPR}, but found that the selected one lead to more stable training.

\paragraph{Confidence and normal map $\rightarrow \confidence, \normapenew$}
In order for \eq{blender} to determine whether a pixel in image $\imagecloud$ contains appropriate information for rendering from viewpoint $v$ we provide two sources of information: a normal map and a confidence score.
The normal map $\normapenew$, processed in a way analogous to $\imagecloud$, can be used to decide whether a pixel in $\imageinput$ has been well observed from the input measurement $\inputview$ (e.g. the network should learn to discard measurements taken at low-grazing angles).
Conversely, the relationship between $\inputview$ and $\viewnew$, encoded by $\confidence$, can be used to infer whether a novel viewpoint is back-facing (i.e. $\confidence<0$) or front-facing it (i.e. $\confidence>0$).
We compute this quantity as the dot product between the cameras view vectors: $\confidence=[0,0,1] \cdot {\mathbf{r}_z}/{\Vert \mathbf{r}_z \Vert}$, where $\inputview$ is always assumed to be the origin and  $\mathbf{r}_z$ is the third column of the rotation matrix for the novel camera viewpoint $\viewnew$.
An example of input and output of this module can be observed in \Figure{framework}, top row.

\subsection{Calibration Image Selector}
In a pre-processing stage, we collect a set of calibration images $\{\imagecalib^n\}$ from the user with associated poses $\{\posecalib^n\}$ .
For example, one could ask the user to rotate in front of the camera before the system starts; an example of calibration set is visualized in the second row of \Figure{framework}.
While it is unreasonable to expect this collection to contain the user in the desired pose, and observed exactly from the viewpoint $\viewnew$, it is assumed the calibration set will contain enough information to extrapolate the appearance of the user from the novel viewpoint $\viewnew$.
Therefore, in this stage we \textit{select} a reasonable image from the calibration set that, when warped by \eq{warper} will provide sufficient information to \eq{blender} to produce the final output. 
We compute a score for all the calibration images, and the calibration image with the highest score is selected.
A few examples of the selection process are shown in the supplementary material.
Our selection score is composed of three terms:
\begin{equation}
S^n =
\omega_\text{head} S_\text{head}^n + 
\omega_\text{torso} S_\text{torso}^n + 
\omega_\text{sim} S_\text{sim}^n
\end{equation}
From the current 3D keypoints $\posenew$, we compute a 3D unit vector representing the forward looking direction of the user's head.
The vector is computed by creating a local coordinate system from the keypoints of the eyes and nose.
Analogously, we compute 3D unit vectors $\{d^n_\text{calib}\}$ from the calibration images keypoints $\{\posecalib^n\}$. 
The head score is then simply the dot product $S^n_\text{head} = d \cdot d^n_\text{calib}$, and a similar process is adopted for $S_\text{torso}^n$, where the coordinate system is created from the left/right shoulder and the left hip keypoints.
These two scores are already sufficient to accurately select a calibration image from the desired novel viewpoint, however they do not take into account the configuration of the limbs.
Therefore we introduce a third term, $S_\text{sim}^n$, that computes a similarity score between the keypoints  $\posecalib^n$ in the calibration images to those in the target pose $\posenew$. 
To simplify the notation, we refer to $\hat\posenew$ and $\hat\pose^n_\text{calib}$ as the image-space 2D coordinates of keypoints in homogeneous coordinates.
We can compute a similarity transformation (rotation, translation, scale) $\mathbf{T}_n\in\R^{3\times3}$ that aligns the two sets.
Note that at least $2$ points are needed to estimate our $4$~DOF transformation (one for rotation, two for translation, and one for scale), therefore we group arm keypoints (elbow, wrist) and leg keypoints (knee, foot) together.
For instance, for all the keypoints belonging to the \textit{left arm} group ($LA$) we calculate: 
\begin{equation}
\arg\min_{\mathbf{T}_{n}^{LA}} \sum_{LA} \| \hat\pose^{LA} - \mathbf{T}_{n}^{LA}\hat\pose^{n,LA}_\text{calib} \|^2
\label{eq:trans}
\end{equation}
We then define the similarity score as:
\begin{equation}
S^{LA} = \exp(-\sigma \Vert \hat\pose - \mathbf{T}_{n}^{LA} \hat\pose^{n,LA}_\text{calib} \Vert )
\end{equation}
The final $S^{n}_{sim}$ is the sum of the scores for the $4$ limbs (indexed by $j$). 
The weights $\omega_j$ are tuned to give more importance to head and torso directions, which define the desired target viewpoint.
The calibration image $\imagecalib$ with the respective pose $\posecalib$ with the highest score $\bar S$ is returned from this stage.
{All the details regarding the chosen parameters can be found in the supplementary material.}

\begin{figure}[t]
\centering
\includegraphics[width=\columnwidth]{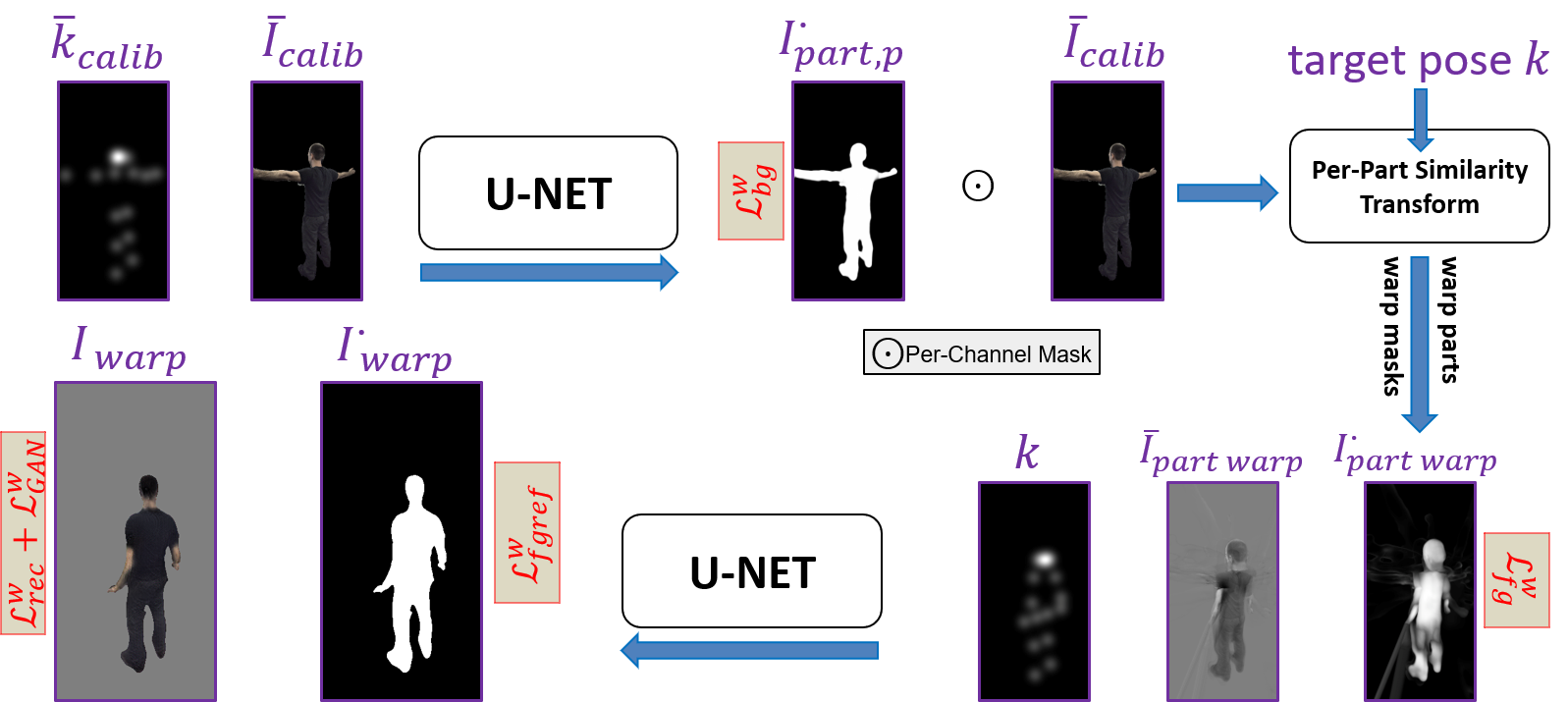}
\caption{The Calibration Warper takes as input the selected calibration the selected calibration image $\imagecalib$ and pose $\posecalib$ and aligns it to the target pose $\posenew$. It also produces a foreground mask $\imagewarpsilho$. For visualization purposes multiple channels are collapsed into a single image. See text for details.}
\label{fig:calib_warper}
\vspace{-15pt}
\end{figure}

\subsection{Calibration Warper}
\label{sec:calib_warper}
The selected calibration image  $\imagecalib$ should have a similar viewpoint to $\viewnew$, but the pose $\posecalib$ could still be different from the desired $\posenew$, as the calibration set is small.
Therefore, we \textit{warp} $\imagecalib$ to obtain an image $\imagewarp$, as well as its silhouette $\imagewarpsilho$.
The architecture we designed is inspired by Balakrishnan~\etal~\cite{balakrishnan_cvpr18}, which uses U-NET modules~\cite{unet}; see \Figure{calib_warper} for an overview.

The calibration pose $\posecalib$ tensor ($17$ channels, one per keypoint) and calibration image $\imagecalib$ go through a U-NET module that produces as output part masks $\{ \bodywarpsilho \}$ plus a background mask $\bgwarpsilho$.
These masks select which regions of the body should be warped according to a similarity transformation.
Similarly to \cite{balakrishnan_cvpr18}, the warping transformations are not learned, but computed via \eq{trans} on keypoint \textit{groups} of at least two 2D points; we have $10$ groups of keypoints (see supplementary material for details).  
The warped texture $\warpedtexture$ has $3$ RGB channels for each keypoints group $p$ ($30$ channels in total).
However, in contrast to  \cite{balakrishnan_cvpr18}, we do not use the masks just to select pixels to be warped, but also warp the body part masks themselves to the target pose $ \posenew$. We then take the maximum across all the channels and supervise the synthesis of the resulting warped silhouette $\warpedmask$.
We noticed that this is crucial to avoid overfitting, and to teach the network to \textit{transfer} the texture from the calibration image to the target view {and keeping high frequency details}.
We also differ from \cite{balakrishnan_cvpr18} in that we do not synthesize the background, as we are only interested in the performer, but we do additionally predict a background mask $\bgwarpsilho$. 

Finally, the $10$ channels encoding the per-part texture $\warpedtexture$ and the warped silhouette mask $\warpedmask$ go through another U-NET module that merges the per-part textures and refines the final foreground mask.
{Please see additional details in the supplementary material.}

The \textit{Calibration Warper} is training minimizing multiple losses:
\begin{align}
\begin{split}
\mathcal{L}_{warp} & = 
w_\text{rec}^\warp \mathcal{L}_\text{rec}^\warp + 
w_\text{fg}^\warp \mathcal{L}_\text{fg}^\warp + 
w_\text{bg}^\warp \mathcal{L}_\text{bg}^\warp + \\ & + 
w_\text{fgref}^\warp \mathcal{L}_\text{fgref}^\warp +
w_\text{GAN}^\warp \mathcal{L}_\text{GAN}^\warp,
\end{split}
\end{align}
where all the weights $w^\warp_*$ are empirically chosen such that all the losses are approximately in the same dynamic range.

\paragraph{Warp reconstruction loss $\mathcal{L}_{rec}^\warp$}
Our perceptual reconstruction loss $\mathcal{L}_{rec}^\warp = \|\text{VGG}(\imagewarp) - \text{VGG}(\imagegt)\|_2$ measures the difference in VGG feature-space between the predicted image $\imagewarp$, and the corresponding groundtruth image $\imagegt$. 
Given the nature of calibration images, $\imagewarp$ may lack high frequency details such as facial expressions. Therefore, we compute the loss selecting features from conv2 up to conv5 layers of the VGG network.

\paragraph{Warp background loss $\mathcal{L}_{bg}^\warp$}
In order to remove the background component of \cite{balakrishnan_cvpr18}, we have a loss $\mathcal{L}_{bg}^\warp = \|\bgwarpsilho - \bgwarpsilhogt\|_1$ between the predicted mask $\bgwarpsilho$ and the groundtruth mask $\bgwarpsilhogt=1-\semanticgt$.
We considered other losses (e.g. logistic) but they all produced very similar results.

\paragraph{Warp foreground loss $\mathcal{L}_{fg}^\warp$}
Each part mask is warped into target pose $\posenew$ by the corresponding similarity transformation.
We then merge all the channels with a max-pooling operator, and retrieve a foreground mask $\warpedmask$, over which we impose our loss $\mathcal{L}_{fg}^\warp = \| \warpedmask - \semanticgt \|_1$. 
This loss is crucial to push the network towards learning transformation rather than memorizing the solution (i.e. overfitting).

\paragraph{Warp foreground refinement loss $\mathcal{L}_{fgref}^\warp$}
The warped part masks $\bodywarpsilho$ may not match the silhouette precisely due to the assumption of similarity transformation among the body parts, therefore we also refine the mask producing a final binary image $ \imagewarpsilho$. This is trained by minimizing the loss  $\mathcal{L}_{fgref}^\warp = \| \imagewarpsilho - \semanticgt \|_1$.

\paragraph{Warp GAN loss $\mathcal{L}_{GAN}^\warp$}
We finally add a GAN component that helps hallucinating realistic high frequency details as shown in \cite{balakrishnan_cvpr18}. Following the original paper \cite{gan} we found more stable results when used the following GAN component: $\mathcal{L}_{GAN}^\warp = -\log(D(\imagewarpsilho))$, where the discriminator $D$ consists of $5$ conv layers with $256$ filters, with max pooling layers to downsample the feature maps. Finally we add $2$ fully connected layers with $256$ features and a sigmoid activation to produce the discriminator label.


\subsection{Neural Blender}
\label{sec:blender}
The re-rendered image $\imagecloud$ can be enhanced by the content in the warped calibration $\imagewarp$ via a neural blending operation consisting of another U-NET module: {please see the supplementary material for more details regarding the architecture}.
By design, this module should always favor details from $\imagecloud$ if the novel camera view $\viewnew$ is close to the original $\inputview$, while it should leverage the texture in $\imagewarp$ for back-facing views.
To guide the network towards this, we pass as input the normal map $\normapenew$, and the confidence $\confidence$, which is passed as an extra channel to each pixel. 
These additional channels contain all the information needed to disambiguate frontal from back views.
The mask $\imagewarpsilho$ acts as an additional feature to guide the network towards understanding where it should hallucinate image content not visible in the re-rendered image $\imagecloud$. 

The \textit{neural blender} is supervised by the following loss:
\begin{align}
\begin{split}
\mathcal{L}_{blender}  = 
w_\text{rec}^\blend \mathcal{L}^\blend_\text{rec} + 
w_\text{GAN}^\blend \mathcal{L}^\blend_\text{GAN} 
\end{split}
\end{align}

\paragraph{Blender reconstruction loss $\mathcal{L}^\blend_\text{rec}$}
The reconstruction loss computes the difference between the final image output $\imageoutput$ and the target view $\imagegt$ . This loss is defined $\mathcal{L}^\blend_\text{rec} = \|\text{VGG}(\imageoutput) - \text{VGG}(\imagegt)\|_2 + w_{\ell_1}\| \imageoutput - \imagegt\|_1$. A small ($w_{\ell_1}=0.01$) photometric ($\ell_1$) loss is needed to ensure faster color convergence.

\paragraph{Blender GAN loss $\mathcal{L}_\text{GAN}^\blend$}
This loss follows the same design of the one described for the calibration warper network.
\begin{figure}[t]
\centering
\includegraphics[width=\columnwidth]{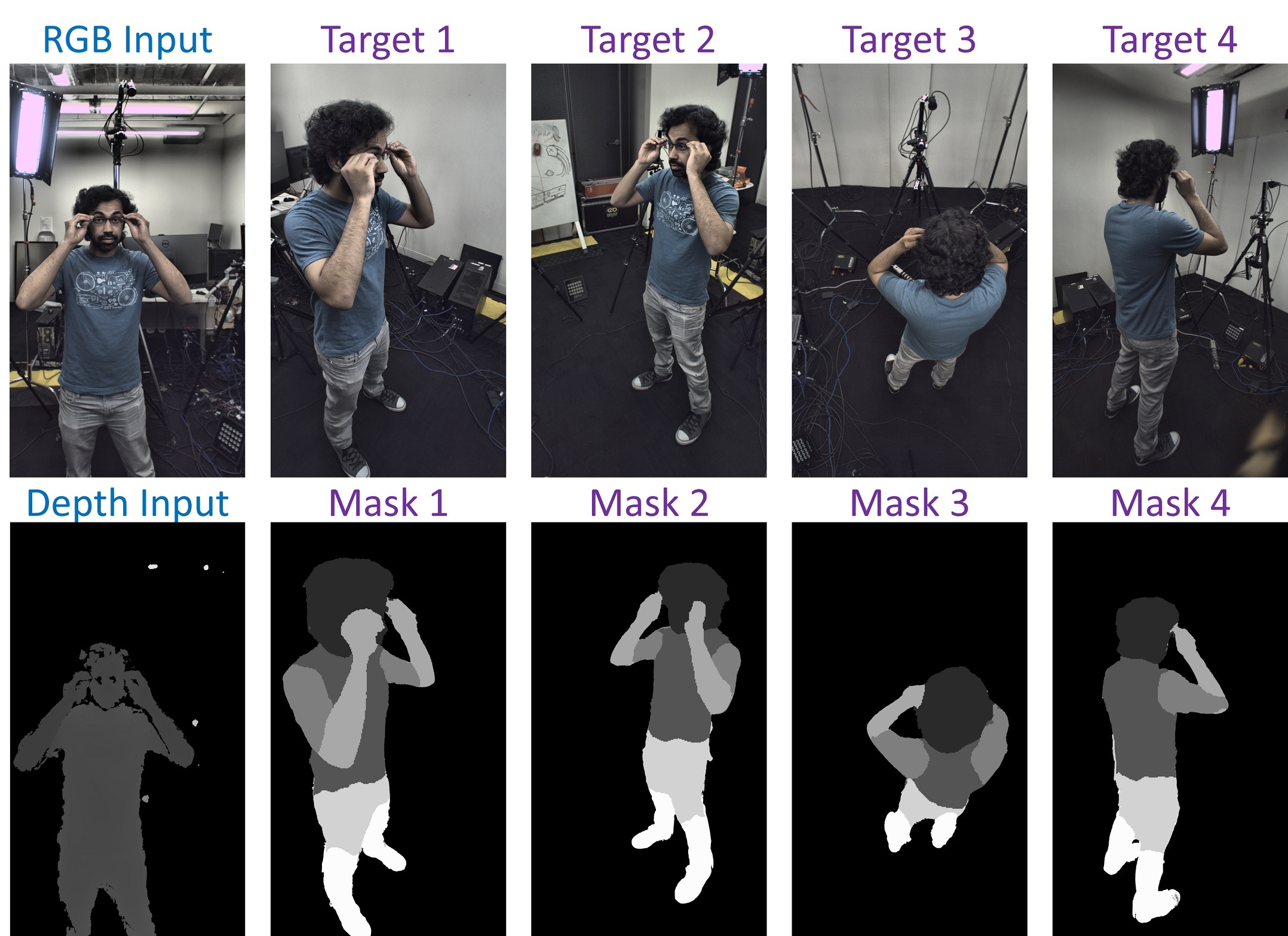}
\caption{Examples of input RGBD and groundtruth novel views with associated masks. Note that in our dataset we have access to $8$ novel views for each input frame.
}
\label{fig:train}
\vspace{-15pt}
\end{figure}

\begin{figure*}[t]
\centering
\includegraphics[width=0.75\linewidth]{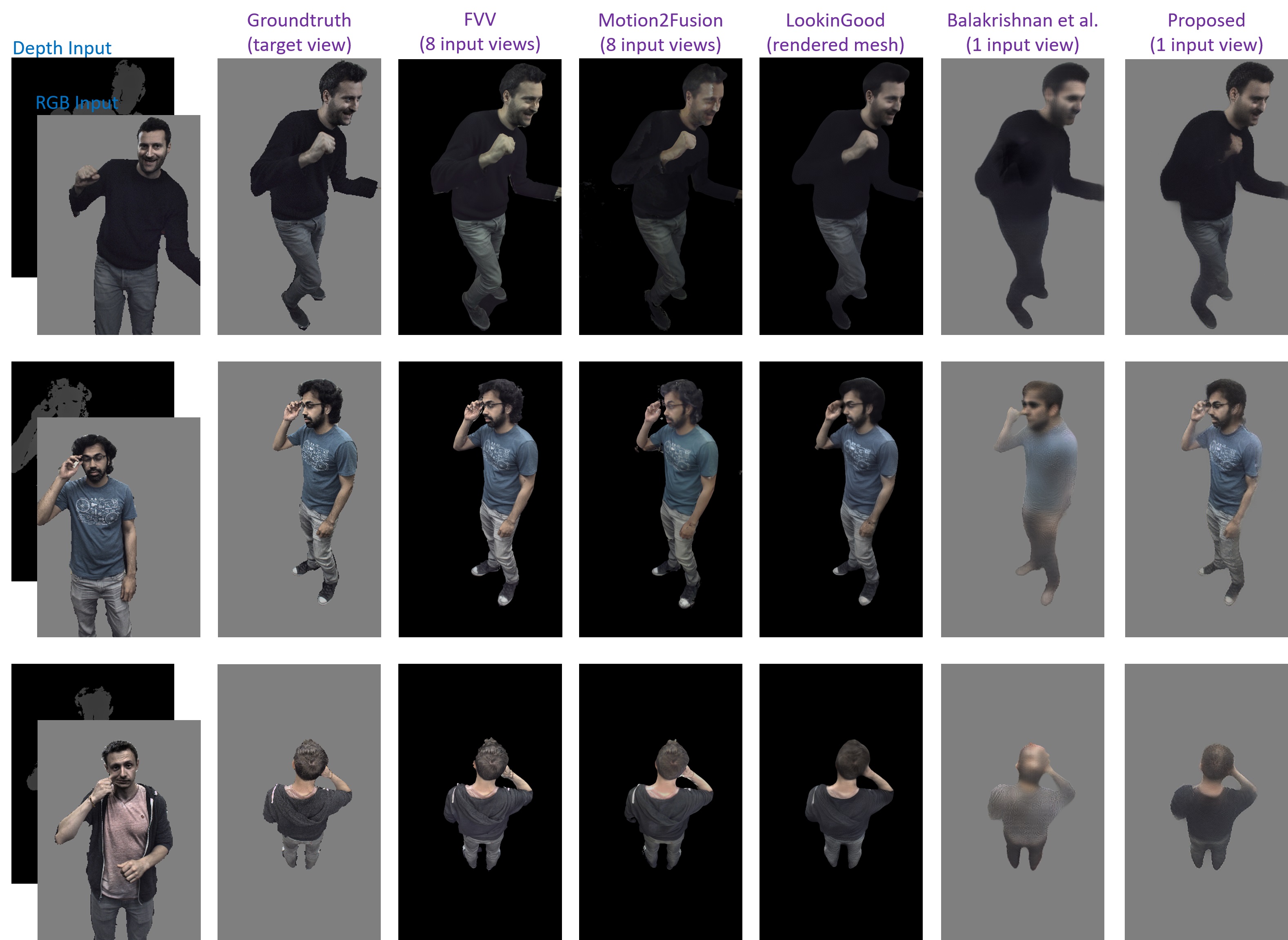}
\caption{Comparisons with state of the art methods. Notice how the proposed framework favorably compares with traditional volumetric capture rigs that use many ($8$) cameras from multiple viewpoints. Notice that due to its real-time nature, Motion2Fusion \cite{dou17} can afford only low resolution ($1280 \times 1024$) RGB images for the texturing phase, whereas FVV \cite{fvv} accepts as input $4000 \times 3000$ images.}
\label{fig:qualitative_sot}
\vspace{-15pt}
\end{figure*}

\section{Evaluation}
\label{sec:evaluation}
We now evaluate our method and compare with representative state-of-the-art algorithms.
We then perform an ablation study on the main components of the system.
All the results here are shown on test sequences not used during training; additional exhaustive evaluations can be found in the supplementary material.
\subsection{Training Data Collection}
\label{sec:dataset}
The training procedure requires input views from an RGBD sensor and multiple groundtruth target views.
Recent multi-view datasets of humans, such as Human 3.6M \cite{h36m_pami}, only provides $4$ RGB views and a \textit{single} low-resolution depth (TOF) sensor, which is insufficient for the task at hand; therefore we collected our own dataset with $20$ subjects.
Similarly to \cite{lookingood}, we used a multi-camera setup with $8$ high resolution RGB views coupled with a custom active depth sensor \cite{sos}. All the cameras were synchronized at at $30$Hz by an external trigger.
The raw RGB resolution is $4000 \times 3000$, whereas the depth resolution is $1280 \times 1024$. 
Due to memory limitations during the training, we downsampled also the RGB images to $1280 \times 1024$ pixels.

Each performer was free to perform any arbitrary movement in the capture space (e.g. walking, jogging, dancing, etc.) while simultaneously performing facial movements and expressions.
For each subject we recorded $10$ sequences of $500$ frames.
For each participant in the training set, we left $2$ sequences out during training.
One sequence is used as calibration, where we randomly pick $10$ frames at each training iteration as calibration images.
The second sequence is used as test to evaluate the performance of a seen actor but unseen actions.
Finally, we left $5$ subjects out from the training datasets to assess the performances of the algorithm on unseen people. 

\paragraph{Silhouette masks generation} As described in \Sec{calib_warper} and \Sec{blender}, our training procedure relies on groundtruth foreground and background masks ($\semanticgt$ and $\bgwarpsilhogt = 1 - \semanticgt$). Thus, we use the state-of-the-art body semantic segmentation algorithm by Chen~\etal~\cite{deeplabv3} to generate these masks $\semanticgt$ which are then refined by a pairwise CRF~\cite{meanfield} to improve the segmentation boundaries. We do not explicit make use of the semantic information extracted by this algorithm such as in \cite{lookingood}, leaving this for future work.
Note that at test time, the segmentation is not required input, but nonetheless we predict a silhouette as a by product as to remove the dependency on the background structure. 
Examples of our training data can be observed in \Figure{train}.
No manual annotation is required hence data collection is fully automatic.

\begin{figure}[t]
\centering
\includegraphics[width=\columnwidth]{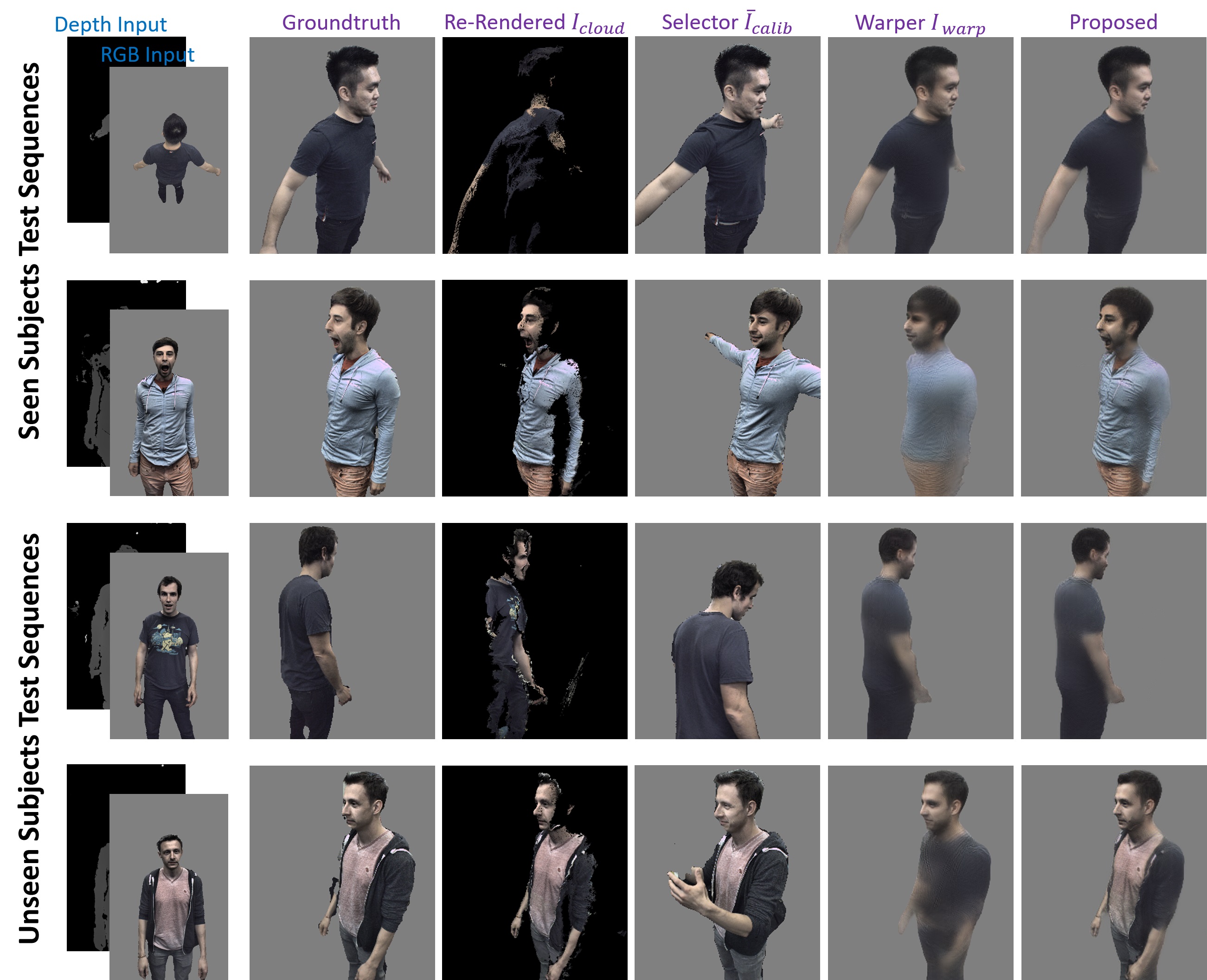}
\caption{Results of the various stage of the pipeline. Notice how each stage of the system contributes to achieve the final high quality results, proving the effectivness of our design choices. Finally, thanks to the semi-parametric model, the algorithm generalizes well across unseen subjects.}
\label{fig:pipeline_results}
\vspace{-10pt}
\end{figure}

\subsection{Comparison with State of the Art}
\begin{table*}[t]
\begin{center}
\begin{tabular}{|  l | c | c | c | c | c || c |c | c|}
\hline
 &  \small{Proposed} & $\imagecloud$ & $\imagecalib$ & $\imagewarp$ &\small{Balakrishnan et al.~\cite{balakrishnan_cvpr18}} & \small{LookinGood ~\cite{lookingood}} & M2F~\cite{dou17} & FVV \cite{fvv} \\ 
  & \small{1 view}  & \small{1 view} & \small{1 view} & \small{1 view} & \small{1 view} & \small{8 views} &  \small{8 views} &  \small{8 views}\\
  \hline
  $\ell_1$ \small{Loss} & $\mathbf{17.40}$ & $27.27$ & $20.02$ & $18.70$ & $18.01$ & $38.80$  & $33.72$  & $\mathbf{7.39}$ \\
  \small{PSNR} & $\mathbf{28.43}$ & $22.35$ & $21.10$ & $27.32$ & $22.93$ & $29.93$ & $28.21$ & $\mathbf{32.60}$\\
  \small{MS-SSIM} & $\mathbf{0.92}$ & $0.84$ & $0.87$ & $0.91$ & $0.86$ & $0.92$ & $\mathbf{0.96}$ & $\mathbf{0.96}$\\
  \small{VGG Loss} & $\mathbf{12.50}$ & $21.20$ & $21.41$ & $13.96$ & $20.16$ & $10.65$ & $\mathbf{5.34}$ & $6.51$\\
\hline
\end{tabular}
\caption{Quantitative evaluations on test sequences. We computed multiple metrics such as Photometric Error ($\ell_1$ loss), PSNR, MS-SSIM and Perceptual Loss. We compared the method with the output of the rendering stage $\imagecloud$, the output of the calibration selector  $\imagecalib$ and the output of the calibration warper  $\imagewarp$. We also show how our method outperforms on multiple metrics the state of the art method of  Balakrishna et al.~\cite{balakrishnan_cvpr18}. We also favorably compare with full capture rig solutions such as Motion2Fusion \cite{dou17}, FVV \cite{fvv} and the LookinGood system \cite{lookingood}.}
\label{tab:res}
\end{center}
\vspace{-15pt}
\end{table*}

We now compare the method with representative state of the art approaches: we selected algorithms for comparison representative of the different strategies they use. The very recent method by Balakrishnan et al.~\cite{balakrishnan_cvpr18} was selected as a state of the art machine learning based approach due to its high quality results. We also re-implemented traditional capture rig solutions such as FVV~\cite{fvv} and Motion2Fusion~\cite{dou17}. Finally we compare with LookinGood~\cite{lookingood}, a hybrid pipeline that combines geometric pipelines with deep networks.
Notice, that these systems use all the available views ($8$ cameras in our dataset) as input, whereas our framework relies on a \textit{single} RGBD view.

\paragraph{Qualitative Results} We show qualitative results on \Figure{qualitative_sot}. Notice how our algorithm, using only a single RGBD input, outperforms the method of Balakrishnan et al.~\cite{balakrishnan_cvpr18}: we synthesize sharper results and also handle viewpoint and scale changes correctly. Additionally, the proposed framework generates compelling results, often comparable to multiview methods such as LookinGood \cite{lookingood}, Motion2Fusion \cite{dou17} or FVV \cite{fvv}.


\paragraph{Quantitative Comparisons} To quantitatively assess and compare the method with the state of the art, we computed multiple metrics using the available groundtruth images. The results are shown in Table \ref{tab:res}. Our system {clearly outperforms} the multiple baselines and compares favorably to state of the art volumetric capture systems that use multiple input views.

\subsection{Ablation Study}
We now quantitatively and qualitatively analyze each each stage of the pipeline. In \Figure{pipeline_results} notice how each stage of the pipeline contributes to achieve the final high quality result.
This proves that each component was carefully designed and needed.
Notice also how we can also generalize to unseen subjects thanks to the semi-parametric approach we proposed.
These excellent results are also confirmed in the quantitative evaluation we reported in Table \ref{tab:res}: note how the output of the full system consistently outperforms the one from the re-rendering ($\imagecloud$), the calibration image selector ($\imagecalib$), and the calibration image warper ($\imagewarp$).
We refer the reader to the supplementary material for more detailed examples.

\begin{figure}[t]
\centering
\includegraphics[width=\columnwidth]{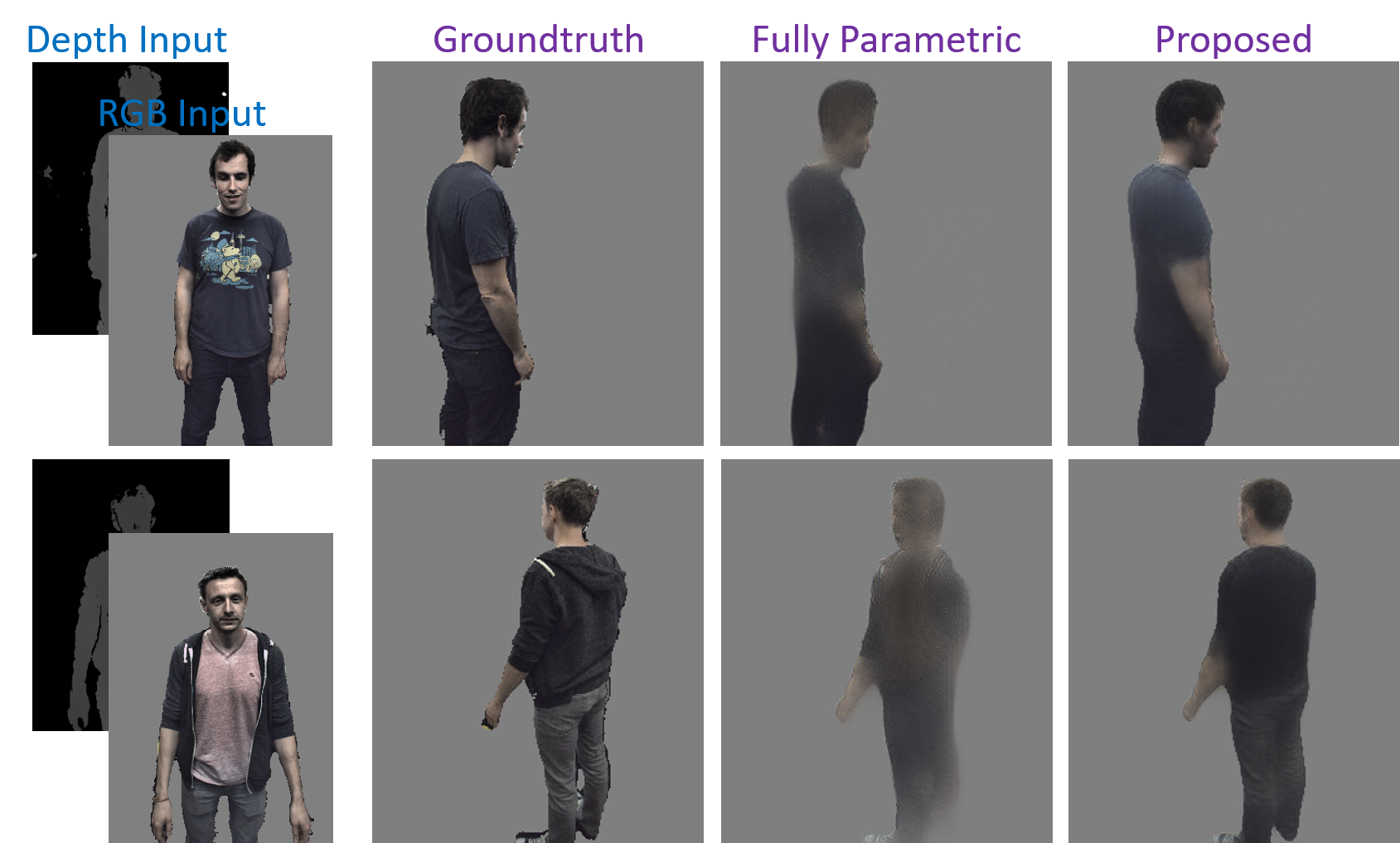}
\caption{Comparison of the proposed system with the fully parametric model. Notice how the semi-parametric part is crucial to get the highest level of quality.}
\label{fig:only_blender}
\vspace{-10pt}
\end{figure}

\paragraph{Comparison with fully parametric model} In this experiment we removed the semi-parametric part of our framework, i.e. the calibration selector and the calibration warper, and train the neural blender on the output of the re-renderer (i.e. a fully parametric model). This is similar to the approach presented in \cite{lookingood}, applied to a single RGBD image. We show the results in \Figure{only_blender}: notice how the proposed semi-parametric model is crucial to properly handle large viewpoint changes.

\begin{figure}[t]
\centering
\includegraphics[width=\columnwidth]{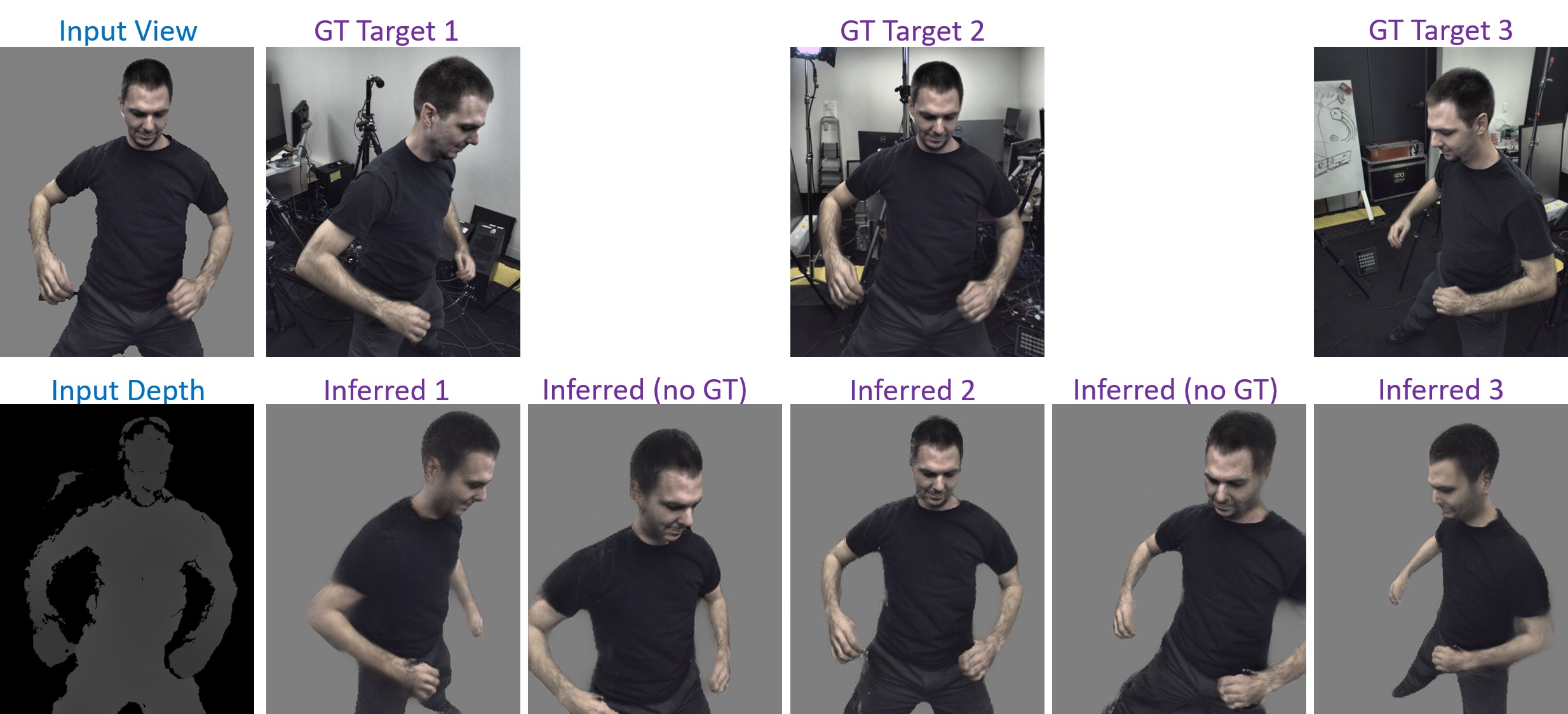}
\caption{Predictions for viewpoints not in the training set. The method correctly infers views where no groundtruth is available.}
\label{fig:viewpoint_invariance}
\vspace{-15pt}
\end{figure}

\paragraph{Viewpoint generalization} We finally show in \Figure{viewpoint_invariance} qualitative examples for viewpoints not present in the training set. Notice how we are able to robustly handle those cases. Please see supplementary materials for more examples.
\section{Conclusions}
We proposed a novel formulation to tackle the problem of volumetric capture of humans with machine learning.
Our pipeline elegantly combines traditional geometry to semi-parametric learning.
We exhaustively tested the framework and compared it with multiple state of the art methods, showing unprecedented results for a single RGBD camera system.
Currently, our main limitations are due to sparse keypoints, which we plan to address by adding additional discriminative priors such as in \cite{totalcapture}.
In future work, we will also investigate performing end to end training of the entire pipeline, including the calibration keyframe selection and warping.

{\small
\bibliographystyle{ieee}
\bibliography{references}
}

\end{document}


\title{Volumetric Capture of Humans with a Single RGBD Camera via Semi-Parametric Learning Supplementary Material}

\author{Rohit Pandey, Anastasia Tkach, Shuoran Yang, Pavel Pidlypenskyi, Jonathan Taylor, \\ Ricardo Martin-Brualla, Andrea Tagliasacchi, George Papandreou, Philip Davidson,  \\ Cem Keskin, Shahram Izadi, Sean Fanello\\
Google Inc.\\
}

\maketitle
\thispagestyle{empty}

In this supplementary material, we provide additional information regarding our method's implementation, more details and more ablation studies on important components of the proposed framework. 

\section{Framework Details}
We detail here the choices of various parameters to aid in reproducing results.

\subsection{Calibration Image Selector - Params}
\begin{figure}[t]
\centering
\includegraphics[width=\columnwidth]{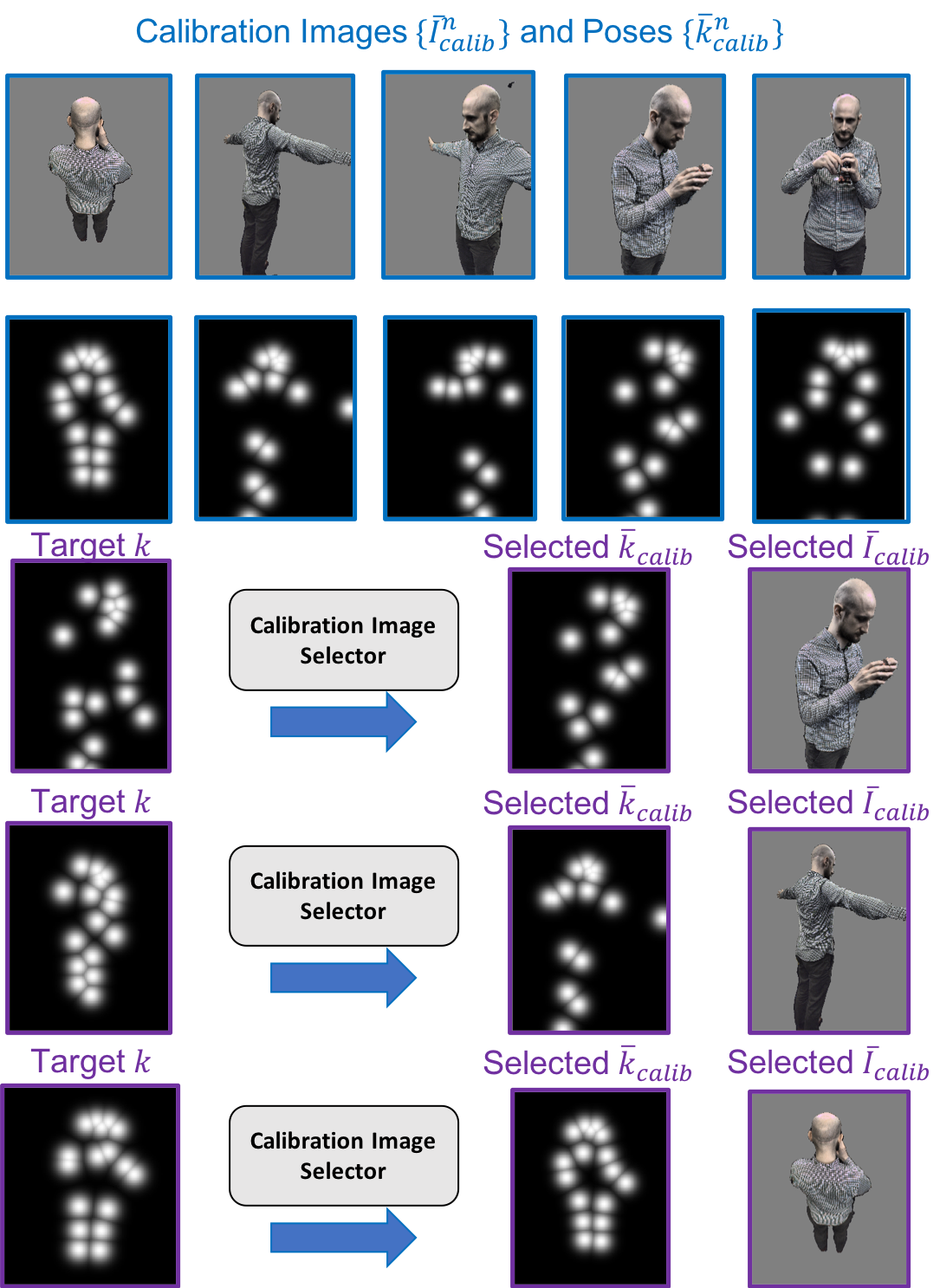}
\caption{
Examples of selected calibration images.
The closest viewpoint to the target is selected.
}
\label{fig:selector_examples}
\vspace{-15pt}
\end{figure}

In Figure \ref{fig:selector_examples} we show some example outputs of the calibration selector module. Note how the module selects the calibration image that most closely matches the viewpoint the person is seen from, based on the target pose. For Eq. $5$ we empirically select $\omega_\text{head}=5$, $\omega_\text{torso}=3$, and $\omega_\text{sim}=1$ so as to weigh the head and torso components of the score highest, then factor in the transformation scores of the limbs.

\subsection{Calibration Image Warper}
\textit{Keypoint grouping}: We detect $17$ keypoints and group them into into $10$ body parts. The body parts consist of 1) head 2) body 3) left upper arm 4) right upper arm 5) left lower arm 6) right lower arm 7) left upper leg 8) right upper leg 9) left lower leg, and 10) right lower leg. The head keypoints consist of the nose, left/right eyes, and left/right ears. The body keypoints consist of the left/right shoulder, and left/right hip keypoints. Each limb consists of two keypoints; the shoulder and elbow for the upper arms, the elbow and wrist for the lower arms, the hip and knee for the upper legs, and the knee and ankle for the lower legs.

\textit{Network architecture}: Our U-Net architecture consists of $5$ encoder blocks followed by $5$ decoder blocks. Each encoder block downsamples the input by a factor of $2$ and consists of $2$ convolutional layers; the first with a kernel size of $3$ and stride $1$ and the second with a kernel size of $4$ and stride $2$. Each decoder block consists of a bilinear upsampling layer that upsamples the input by a factor of $2$, followed by a convolutional layer with kernel size $3$ and stride $1$. The encoder blocks use $64$ filters for the first block followed by $128$ filters for the remaining $4$ blocks. The decoder blocks use $128$ filters for the first $4$ convolutional layers and $32$ filters for the final block. Additionally, we add skip connections from the encoder to the decoder in the form of concatenation of feature maps of matching size. Leaky ReLu activations (with $alpha=0.2$) are used for all convolutional layers.

The calibration image warper module adds a final convolutional layer with $4$ channels to produce the RGB output $\imagewarp$, and the mask $\imagewarpsilho$. Tanh activation is used for the RGB and sigmoid for the mask prediction.

\subsection{Neural Blender}
\begin{figure}[t]
\centering
\includegraphics[width=\columnwidth]{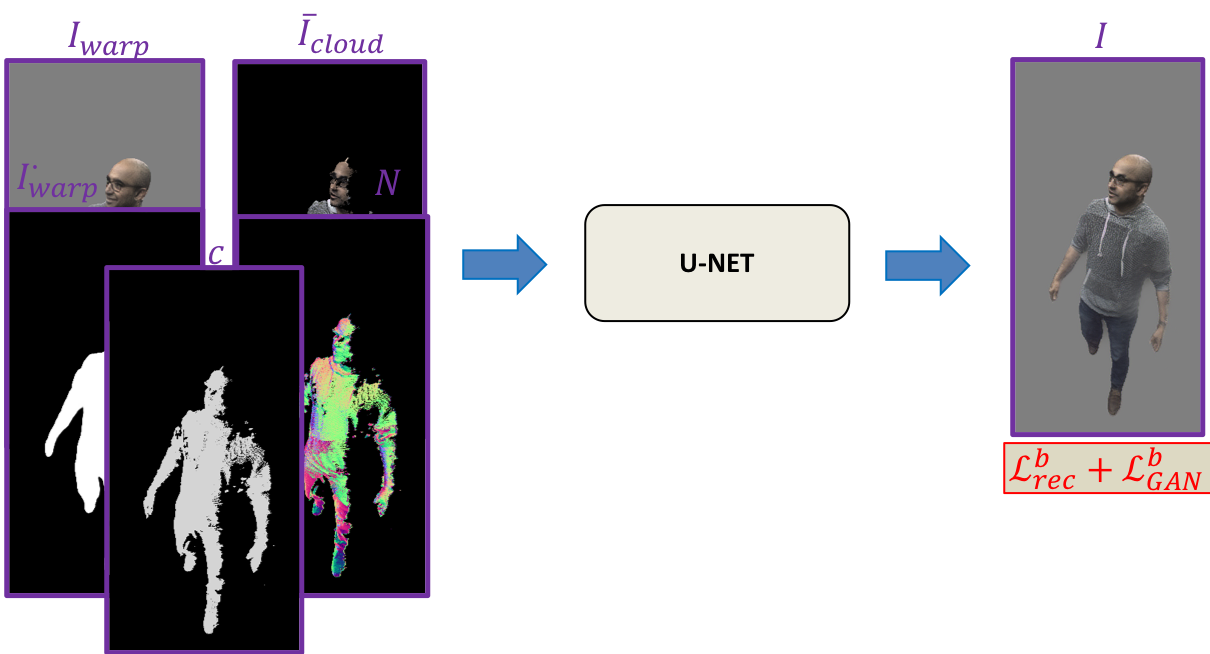}
\caption{The final Neural Blender module takes the output RGB and mask of the calibration warper module, as well as the warped RGB, normals, and viewpoint confidence from the re-rendering module, and learns how to blend them into the final output RGB.
}
\label{fig:blender}
\vspace{-15pt}
\end{figure}

The neural blender module is shown in Figure \ref{fig:blender}. For simplicity we re-use the same U-Net architecture described in the calibration image warper module section above. However, the last convolutional layer now outputs only $3$ channels for the final blended RGB.

\subsection{Training Details}
Our networks are implemented in Tensorflow and trained in parallel on 16 NVIDIA V100 GPUs each with $16$ GB of memory. We use the Adam optimizer~\cite{adam} with a learning rate of $1^{-5}$ for the generator and $1^{-6}$ for the discriminator. We perform light data augmentation during training with random cropping in a size range of $0.85$ to $1.$ times of the input image size. Additionally, we add standard $\ell_2$ loss regularization (with weight $1^{-5}$) to the weights of the network. We found that our data augmentation and regularization, coupled with the variations introduced via using all possible combinations of source and target cameras in our training set, were sufficient to prevent over-fitting and make our network generalize to unseen poses, viewpoints and people.

\section{Running time of the system}
The proposed architecture has an end-to-end runtime of $104.3 ms$ on a Titan V GPU. Note that, this is the unoptimized runtime and does not take advantage of float16 inference or tensor cores on the volta architectures. We leave achieving realtime inference using the proposed architecture for future work.

\section{Additional Evaluation}
\begin{figure}[t]
\centering
\includegraphics[width=0.80\columnwidth]{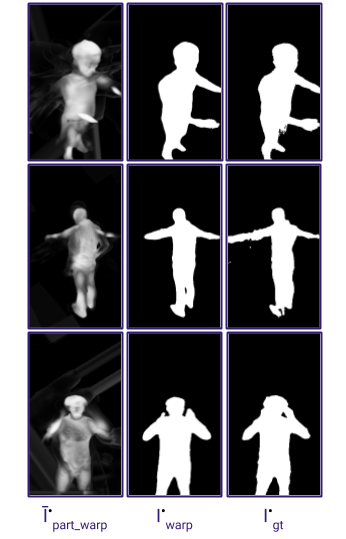}
\caption{Examples of the warped foreground mask and refined foreground mask, compared to the ground truth mask.}
\label{fig:mask_prediction}
\vspace{-10pt}
\end{figure}
In Figure \ref{fig:mask_prediction} we show some examples of the predicted warped part masks and refined masks compared to the ground truth foreground mask. Note that the warped part mask is limited in the accuracy of the silhouette it can produce due to the assumption of 2D similarity transformation between body parts, however, the predicted refined mask is able to overcome these limitations and produce a much cleaner silhouette.

\begin{figure}[t]
\centering
\includegraphics[width=\columnwidth]{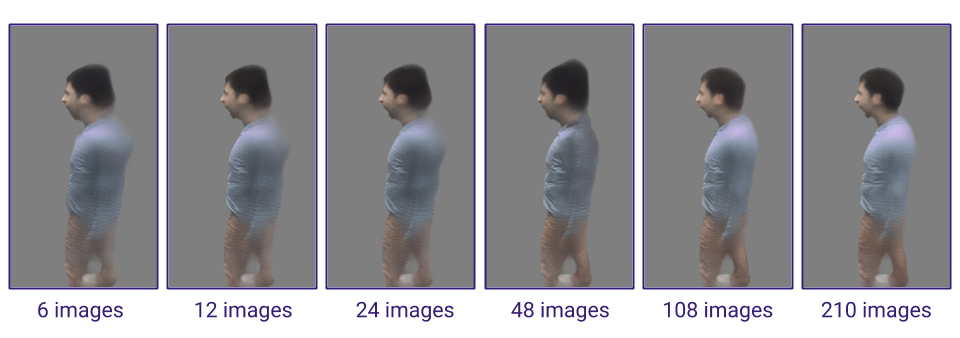}
\caption{Effect of the number of calibration images in the pool on the output of the calibration image warper module.}
\label{fig:num_calib_images}
\vspace{-10pt}
\end{figure}
In Figure \ref{fig:num_calib_images} we show the effect of adding more images to the calibration image pool on the output of the calibration warping module. All calibration images are chosen at random from a held out sequence of the user. Notice how the quality of the output improves as the number of calibration images increases from $6$ to $210$. This is due to the higher probability of finding a calibration image with matching viewpoint as we increase the size of the image pool.

\begin{figure}[t]
\centering
\includegraphics[width=\columnwidth]{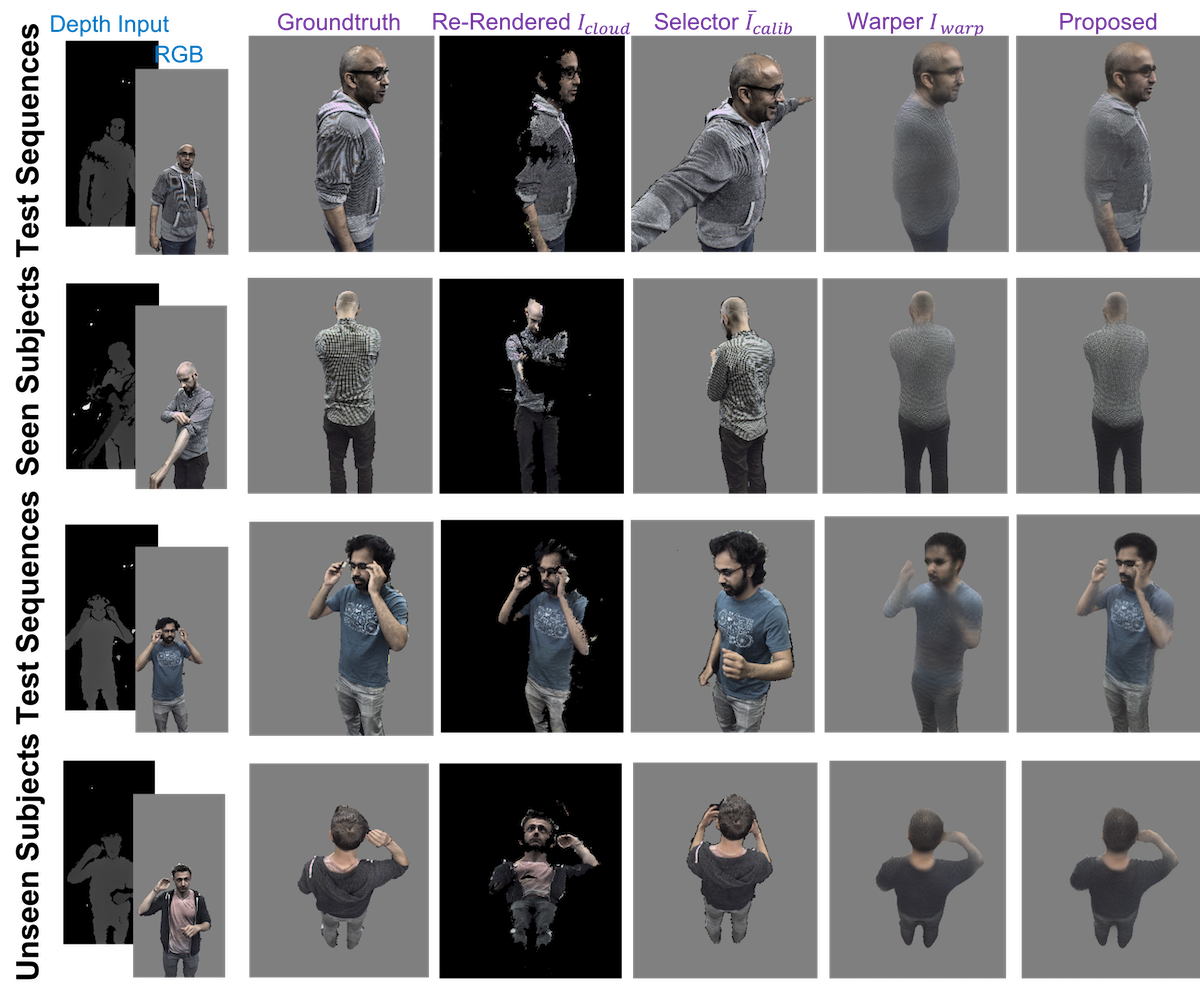}
\caption{Additional results showing various stages of the pipeline.}
\label{fig:supp_pipeline_results}
\vspace{-10pt}
\end{figure}

In Figure \ref{fig:supp_pipeline_results} we present additional results showing the output of various stages of the proposed pipeline on seen and unseen subjects.

\begin{figure}[t]
\centering
\includegraphics[width=\columnwidth]{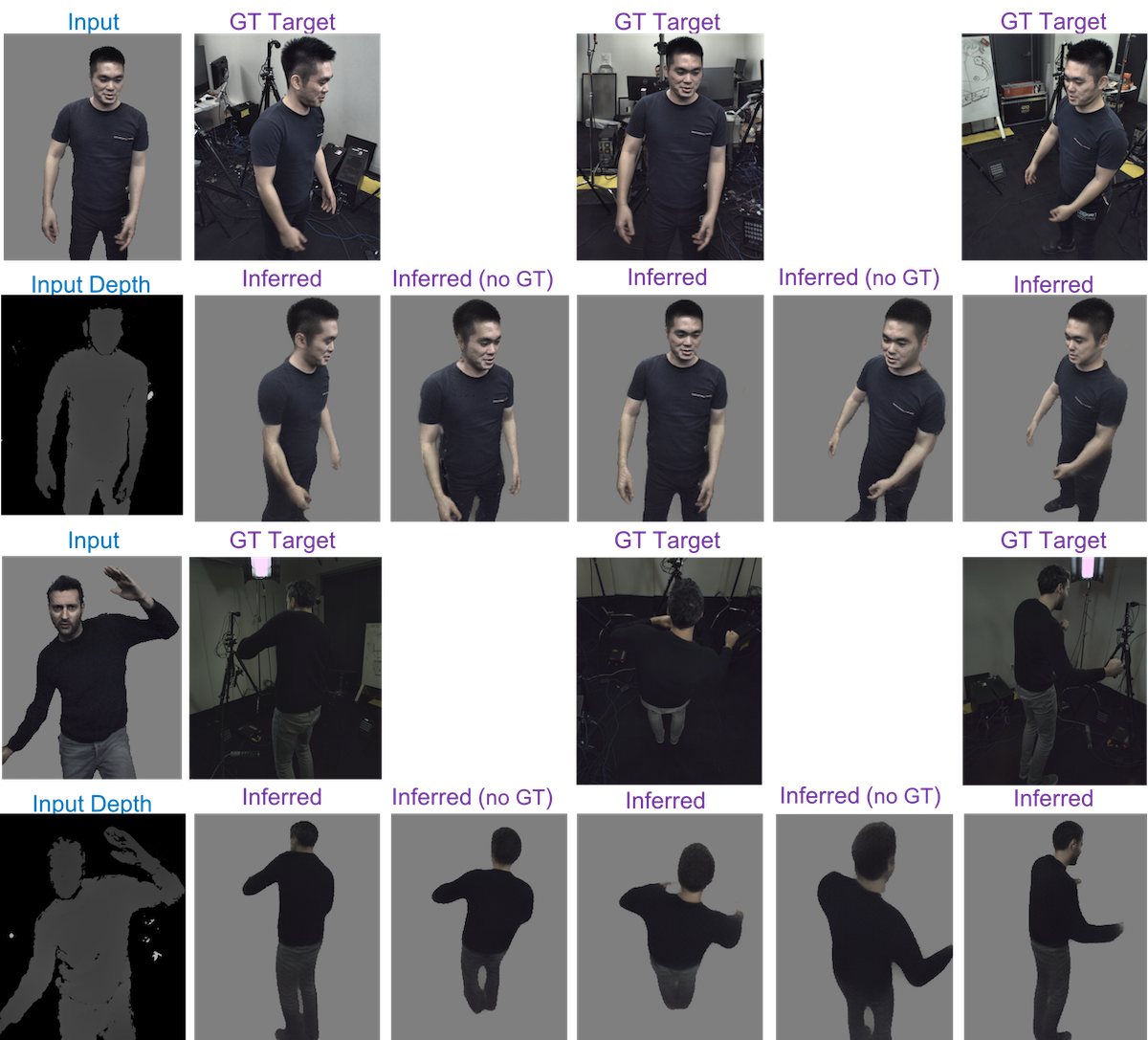}
\caption{Additional results showing the viewpoint generalization of the proposed method.}
\label{fig:supp_viewpoint_invariance}
\vspace{-15pt}
\end{figure}

In Figure \ref{fig:supp_viewpoint_invariance} we present additional results showing the ability of the proposed method to generalize to viewpoints not in the training data.

\section{Limitations and Future Work}
One of the limitations of the proposed approach is that the calibration warper produces blurry results when the viewpoint of the selected calibration image is far from the target viewpoint. We notice this in results when the calibration image pool is small as shown in Figure \ref{fig:num_calib_images}. A larger calibration pool or a predefined calibration sequence where the user turns around the camera can help alleviate this issue. Another limitation of the system is that it struggles to produce reasonable outputs where keypoints are not present, for example hands. We hypothesize that adding additional finger keypoints like fingers and more facial keypoints can help both the calibration selection module, as well as the hallucination modules, to produce better results. Finally, the system shows some temporal flickering as can be seen in the supplementary video. This is especially evident when the selected calibration image changes, and likely can be alleviated via temporal architectures like RNNs or temporal coherency losses.

{\small
\bibliographystyle{ieee}
\bibliography{references}
}